\documentclass[USenglish]{article}

\usepackage{xcolor}
\usepackage{amsthm}
\usepackage[small]{dgruyter}

\newcommand{\norm}[1]{\left\lVert#1\right\rVert}
\newcommand{\innerpro}[1]{\left\langle#1\right\rangle}
\newcommand{\bX}{\mathbf{X}}
\newcommand{\bY}{\mathbf{Y}}
\newcommand{\bz}{\mathbf{z}}
\newcommand{\lam}{\lambda}

\newcommand{\x}{{\bf x}}
\newcommand{\hk}{{\mathcal{H}_K}}
\newcommand{\ie}{i.e. }

\newtheorem{theorem}{Theorem}[section]
\newtheorem{remark}{Remark}
\newtheorem{lemma}[theorem]{Lemma}
\newtheorem{assumption}[theorem]{Assumption}

\DeclareSymbolFont{bbm}{U}{msb}{m}{n}
\DeclareSymbolFontAlphabet{\mathbbm}{bbm}

\def\R{\mathbbm R}

\usepackage[utf8]{inputenc}%(only for the pdftex engine)
\usepackage{microtype}

\begin{document}

  \articletype{Chapter}

  \author[1]{Duc Hoan Nguyen}
  \author[1]{Sergei Pereverzyev}
  \author[1,2]{Werner Zellinger} 
  \runningauthor{DC Nguyen, S Pereverzyev and W Zellinger}
  \affil[1]{Johann Radon Institute for Computational and Applied Mathematics, Austrian Academy of Sciences \quad\quad\quad\quad\quad\quad\quad\quad\quad
  \textbf{Contact}, werner.zellinger@ricam.oeaw.ac.at}
  % \affil[2]{contact \textit{werner.zellinger@ricam.oeaw.ac.at}}
  \title{General regularization in covariate shift adaptation}
  \runningtitle{Regularized covariate shift adaptation}
  % \subtitle{...}
  \abstract{
  Sample reweighting is one of the most widely used methods for correcting the error of least squares learning algorithms in reproducing kernel Hilbert spaces (RKHS), that is caused by future data distributions that are different from the training data distribution.
  In practical situations, the sample weights are determined by values of the estimated Radon-Nikod\'ym derivative, of the future data distribution w.r.t.~the training data distribution.
  In this work, we review known error bounds for reweighted kernel regression in RKHS and obtain, by combination, novel results.
  We show under weak smoothness conditions, that the amount of samples, needed to achieve the same order of accuracy as in the standard supervised learning without differences in data distributions, is smaller than proven by state-of-the-art analyses.
  }
  % \classification[PACS]{...}
  % \communicated{...}
  % \dedication{...}
  % \received{...}
  % \accepted{...}
  % \journalname{...}
  % \journalyear{...}
  % \journalvolume{..}
  % \journalissue{..}
  % \startpage{1}
  % \aop
  % \DOI{...}

\maketitle

\section{Introduction} 

%\paragraph*{Learning Problem}
Over the past few decades, data-based algorithms have resulted in significant advances in an extensive variety of different fields. Nevertheless, a noticeable disparity has emerged between the theoretical assumptions that form the basis of algorithmic development and the practical conditions in which these algorithms are deployed. In learning theory, one studies the relationship between the explanatory (input) variable $x \in \mathbf{X} \subset \R^{d_1}$ and the response (output) variable $y \in \mathbf{Y} \subset \R^{d_2}$ under the assumption that they are governed by an unknown probability measure $p(x,y)$ on $\bX \times \mathbf{Y}$. This means that an input $x \in \mathbf{X}$ does not determine uniquely an output  $y \in \mathbf{Y}$, but rather a conditional probability $\rho (y|x)$ of $y$ given $x$, which is assumed to be unknown. Then one uses a training data sample $\bz = \{(x_i,y_i), x_i \in \mathbf{X}, y_i \in \mathbf{Y}, i = 1,2,\ldots, n\}, |\bz| = n,$ drawn independently and identically (i.i.d) from the measure $p(x,y)$ to infer a function $f: \bX \to \mathbf{Y}$ which predicts the label $y' \in \mathbf{Y} $ of any future input $x' \in \bX$ \cite{Cucker2001OnTM,vapnik2013nature}. This problem is an inverse problem, see e.g.~\cite{de2005learning,smale2007,bauer2007}. The error of the prediction model $f$ is quantified by the expected risk 
\begin{align}\label{eq:generalization_error}
    \mathcal{R}_p(f)=\int_{\bX \times \bY} \ell (f(x),y) d p(x,y)
\end{align}
for some \textit{loss} function $\ell:\bY\times\bY\to[0,\infty)$, e.g., the squared loss $\ell(f(x),y)=(f(x)-y)^2$. The choice of the expected risk $\mathcal{R}_p (f)$ realizes a core assumption of learning theory:
\begin{align*}
    \text{\it{Any new example $(x',y')$ is drawn from the same probability measure $p$}}.
\end{align*}
This assumption is, however, often violated in practice.
For example, in medical image analysis, pre-trained learning models, are applied to data from patients following distributions that are different from the training one~\cite{meyer2021domain}.
Chemical measurement systems need to be re-calibrated for new distributions after changes in system setups~\cite{nikzad2018domain}.

%\paragraph*{Covariate Shift Adaptation}
To overcome this problem, unsupervised domain adaptation arises as a relevant approach when the underlying relationship between input samples $x \in \mathbf{X}$ and their corresponding outputs $y \in \mathbf{Y}$ is not exclusively governed by a single probability measure $p(x,y)$.  Instead, it encompasses the situation where %multiple probability measures, 
one more probability measure, denoted as $q(x,y)$, appears to characterize the joint distribution over $\mathbf{X} \times \mathbf{Y}$. 
%This means that an input $x \in \mathbf{X}$ does not determine uniquely an output  $y \in \mathbf{Y}$, but rather a conditional probability $\rho (y|x)$ of $y$ given $x$, which is assumed to be unknown. 
In contrast to the classical problem of learning from examples, in domain adaptation one uses the training data sample $\bz$ drawn independently and identically (i.i.d) from one of the measures, say $p(x,y)$, to reduce the expected risk of the prediction model $f$ over the other measure $q(x,y)$. In the context of domain adaptation, $p(x,y)$ and $q(x,y)$ are called, respectively, the \textit{source} probability and the \textit{target} probability.

\paragraph*{Covariate shift assumption}
In general, the domain adaptation problem with different source and target probabilities is unsolvable, as $p(x,y),$ $q(x,y)$ could be arbitrarily far apart. Therefore, in the present study, we follow \cite{shimodaira2000}, \cite{huang2006} and rely on the so-called covariate shift assumption, where only probabilities of inputs in the source (S) and the target (T) domains (marginal probabilities) $\rho_S (x)$ and $\rho_T (x)$ differs, while the conditional probability $\rho (y|x)$ is the same under both the source and the target probabilities. This means that the joint probabilities $p(x,y)$, $q(x,y)$ can be factorized as the following products
	\begin{align}\label{covar_shift_assump}
	p(x,y) = \rho(y|x)\rho_S (x),\text{       } q(x,y) = \rho(y|x)\rho_T (x).
	\end{align}
In this work, we restrict ourselves to learning with least squares loss, where the expected risk of the prediction of $y$ from $x$ by means of a function $f: \mathbf{X} \rightarrow \mathbf{Y}$ is defined in the target domain as 
	$$\mathcal{R}_q (f) := \int_{\mathbf{X} \times \mathbf{Y}} (f(x) - y)^2 d q(x,y).$$ 
It is easy to check that $\mathcal{R}_q (f)$ attains its minimum at the so-called regression function
	\begin{align}\label{regress_func}
	f(x) = f_{q} (x) = \int_{\mathbf{Y}} y d \rho(y|x )
	\end{align}
see, e.g.,~\cite[Proposition~1]{Cucker2001OnTM}.
However, in the unsupervised domain adaptation setting, neither $\mathcal{R}_q (f)$ nor $f_{q} (x)$ can be computed, because the information about underlying probability $q(x,y)$ is only provided in the form of a set $\x' = (x'_1,x'_2,\ldots,x'_m), |\x'| = m$, of unlabeled examples $x'_i$ of inputs drawn i.i.d. from the target marginal probability measure $\rho_T (x)$.

\paragraph*{Goal}
The goal of unsupervised domain adaptation is to use this information, together with training data $\bz$, to approximate the ideal minimizer $f_q$ by an empirical estimator $f_\bz$ in the sense of excess risk
	$$\mathcal{R}_q (f_\bz) - \mathcal{R}_q (f_q) = \norm{f_\bz - f_q}^2_{L_{2,\rho_T}};$$
	here $L_{2,\rho_T}$ is the space of square integrable functions $f: \mathbf{X} \rightarrow \mathbb{R}$ with respect to the marginal probability measure $\rho_T$.

%\paragraph*{Research Questions}
It can be observed that under covariate shift assumption (\ref{covar_shift_assump}), the expected risks $\mathcal{R}_p (f)$ in supervised learning and $\mathcal{R}_q (f)$ in domain adaptation attain their minimum at the same regression function $f^* (x) = f_p (x) = f_q(x)$ given by (\ref{regress_func}).
Therefore, in unsupervised domain adaptation under covariate shift the aim of approximation is the same as in the standard supervised learning, and it is logical to adjust the methods developed there to the domain adaptation scenario.

One natural step towards this direction is the combination of sample reweighting with regularized least squares regression; a procedure generally referred to as importance weighted regularized least squares (IWRLS)~\cite{shimodaira2000, sugiyama2005, kanamori2009}.
However, although IWRLS is one of the major approaches to unsupervised domain adaptation, its analysis is still in its early developments.
Even for learning with least-squares loss in reproducing kernel Hilbert spaces (RKHS), one of the most well understood directions in statistical learning theory, risk bounds have been developed only recently.

\paragraph*{Contribution}
In this work, we discuss recent results for the analysis of IWRLS in RKHS under the covariate shift assumption.
Our main focus is on general regularization schemes, often referred to as spectral regularization (cf.~\cite{bauer2007}, \cite{rosasco2008}, \cite{lu2018}, and the references therein).
As a result, we refine state-of-the-art risk bounds for IWRLS by combining known results.
In particular, we show how smoothness conditions for Radon-Nikod\'ym differentiation allow IWRLS to achieve the same order of accuracy as regularized kernel regression, but with a smaller amount of samples as in known situations.

In Section~\ref{sec2}, we describe the IWRLS algorithm and review the recent risk bound~\cite{gizewski2022regularization}.
To the best of our knowledge, this is the first risk bound for IWRLS.
This analysis is based on smoothness conditions on $f_q$ and it assumes access to the values of the (unknown) Radon-Nikod\'ym derivative $\frac{d \rho_T}{d \rho_S}$ of $\rho_T$ w.r.t.~$\rho_S$, which appear as the sample weights in IWRLS.

In Section~\ref{sec:radon_Nikod\'ym_differentiation}, we review recent error bounds of~\cite{nguyen2023regularized} for algorithms estimating the Radon-Nikod\'ym derivative which refine the study~\cite{gizewski2022regularization}.
In particular, we highlight general smoothness conditions, under which upper bounds on the pointwise error are smaller than upper bounds on the error in RKHS norm.

In Section~\ref{sec:embedded_regularization}, we embed the regularized Radon-Nikod\'ym differentiation considered in Section~\ref{sec:radon_Nikod\'ym_differentiation} into the regularization schemes considered in Section~\ref{sec2}, so that no exact values of the unknown Radon-Nikod\'ym derivative are needed.
The considered smoothness conditions provide novel situations under which IWRLS achieves the same order of accuracy as standard least squares regression in RKHS.
Interestingly, the obtained order of accuracy is much higher than anticipated under the slightly weaker conditions of the state of the art~\cite{gizewski2022regularization}.

In Section~\ref{sec:parameter_choice}, we review a general method for regularization parameter choice issues in IWRLS~\cite{gizewski2022regularization,dinu2022aggregation}.
The method is based on aggregating several regularized solutions.
Again, the considered smoothness conditions allow to refine known error bounds for this method.

\section{Importance weighted regularized least squares}
\label{sec2}

In the following, we summarize recent results from~\cite{gizewski2022regularization}, which are, to the best of our knowledge, the first risk bounds for IWRLS.
Leter, in Section~\ref{sec:embedded_regularization}, we will refine these bounds.

Since we have no direct access to the target probability measure $  \rho_T  $ and to the space $     {L_{2,\rho_T}}  $ in which we are going to approximate the regression function $f^* = f_q,$ some additional assumptions should be imposed on the relationship between the source probability $ \rho_S $ and the target probability $  \rho_T  $.
In the present study we follow \cite{huang2006} and assume that there is a function $\beta: \mathbf{X} \rightarrow \R_+$ such that 
	\begin{align*}
	d\rho_T (x) = \beta (x)  d\rho_S (x).
	\end{align*}
Then $\beta (x)$ can be viewed as the Radon-Nikod\'ym derivative $\frac{d\rho_T}{d\rho_S}$ of the target measure with respect to the source measure.

In this section, we assume that we have access to the values $\beta_i:=\beta(x_i)$ of the Radon-Nikod\'ym derivative $\beta (x) = \frac{d \rho_T (x)}{d\rho_S (x)} $ at the points $x_i, i = 1,2,\ldots,n$, drawn i.i.d from $\rho_S (x)$.
Moreover, we assume that for any $x \in \mathbf{X}$, $| \beta(x)| \leq b_0$ for some $b_0 > 0 $, as in \cite{huang2006}. Let $\mathcal{H}_K$ be a reproducing Kernel Hilbert space with a positive-definite function $K: \bX \times \bX \rightarrow \R$ as reproducing kernel. We assume that $K$ is a continuous and bounded function, such that for any $x \in \mathbf{X} $ $$\norm{K(\cdot,x)}_{\mathcal{H}_K} = \innerpro{K(\cdot,x), K(\cdot,x)}_{\mathcal{H}_K}^\frac{1}{2} = [K(x,x)]^\frac{1}{2} \leq \kappa_0 < \infty. $$
Recall that the information about the source and target marginal measures are only provided in the form of samples $\x = \{x_1, x_2,\ldots,x_n\} $ and $\x' = \{x_1', x_2',\ldots,x_m'\}$ drawn independently and identically (i.i.d) from $\rho_S$ and $\rho_T$ respectively. In the sequel, we distinguish two sample operators
\begin{align*}
&S_{\x'} f = (f(x_1'), f(x_2'),\ldots,f(x_m')) \in \R^m,\\
&S_{\x} f = (f(x_1), f(x_2),\ldots,f(x_n)) \in \R^n,
\end{align*}
acting from $\mathcal{H}_K$ to $\R^m$ and $\R^n$, where the norms in later spaces are generated by $m^{-1}$-times and $n^{-1}$-times the standard Euclidean inner products. Then the adjoint operators $S_{\x'}^*: \R^m \rightarrow \mathcal{H}_K$ and 
$S_{\x}^*: \R^n \rightarrow \mathcal{H}_K$ are given as
\begin{align*}
&S_{\x'}^* u(\cdot) = \frac{1}{m} \sum_{j = 1}^m K(\cdot, x_j') u_j, \hspace*{0.5cm} u = (u_1,u_2,\ldots,u_m) \in \R^m, \\
&S_{\x}^* v(\cdot) = \frac{1}{n} \sum_{i = 1}^n K(\cdot, x_i) v_i,\hspace*{0.5cm} v = (v_1,v_2,\ldots,v_n) \in \R^n.
\end{align*}

In the context of domain adaptation with covariate shift, the objective is to construct an approximation of the minimizer $f^* = f_q$ for the target expected risk $\mathcal{R}_q(f)$ utilizing the available data $\mathbf{z} = \{(x_i,y_i)\}_{i=1}^n$ sampled from the source distribution $p(x,y)$. To accomplish this objective, one popular employed approach is penalized least squares regression combined with sample reweighting, which is also called importance weighted regularized least squares (IWRLS), see, e.g \cite{evgeniou2000}, \cite{sugiyama2005}, and \cite{kanamori2009}. 
%We assume that the approximation belongs to 
If we are looking for approximations in RKHS $\mathcal{H}_K$, then within IWRLS-approach the approximant $f_\mathbf{z} = f_\mathbf{z}^\lambda $ of $f^* = f_q$ is constructed as the minimizer of weighted and penalized empirical risk
$$ \mathcal{R}_{\mathbf{z},\lambda,\beta} (f) = \frac{1}{n} \sum_{i=1}^{n} \beta_i \left( f(x_i) - y_i \right)^2 + \lambda \| f\|^2_{\mathcal{H}_K}.$$
Since $\beta_i$ are assumed to be non-negative, $ \mathcal{R}_{\mathbf{z},\lambda,\beta}  $ can be written in the form of the so-called Tikhonov regularization functional
$$ \mathcal{R}_{\mathbf{z},\lambda,\beta} (f) =\norm{B^{\frac{1}{2}} S_{\x} f - B^{\frac{1}{2}} {\bf y} }^2_{\R^n} + \lambda \| f\|^2_{\mathcal{H}_K}, $$ where $B^{\frac{1}{2}} = \mathrm{diag}(\sqrt{\beta_1},\sqrt{\beta_2},\ldots,\sqrt{\beta_n}),$ and ${\bf y} = (y_1,y_2,\ldots,y_n)$. The minimizer of $\mathcal{R}_{\mathbf{z},\lambda,\beta}$ admits the following representation 
\begin{align}\label{A.eq2.1.2}
f_\mathbf{z}^\lambda = (\lambda \mathbf{I} + S^*_{{\x}}BS_{{\x}})^{-1} S^*_{{\x}} B {\bf y},
\end{align}
%It is important to note that the operator
and can be seen as an approximate solution to the normal equation $S_{S_X}^* B S_{S_X}f = S^*_{{\x}} B {\bf y}$ regularized by perturbation $\lambda \mathbf{I}f$.
%is a self-adjoint, non-negative, and compact operator on RKHS $\mathcal{H}_K$. Therefore, 
At the same time, the whole arsenal of regularization schemes can potentially be applied to that equation 
%used 
to construct approximations $f_\mathbf{z} = f_\mathbf{z}^\lambda$ of the minimizer $f^* = f_q$ of the target expected risk $\mathcal{R}_q (f)$
%in the form (\ref{A.eq2.1.5}) 
from the data ${\bf z}=({\x}, {\bf y})$ that are sampled from the source measure $p$. In particular, we will use a general regularization scheme to construct a family of 
%IWRLS-approximant (\ref{A.eq2.1.2}) 
approximants as follows
\begin{align}\label{A.eq2.1.5}
f_\mathbf{z}^\lambda = g_\lambda (S_{{\x}}^* B S_{{\x}} ) S_{{\x}}^* B {\bf y},
\end{align}
where $\{g_\lam\}$ is a family of operator functions parametrized by a regularization parameter $\lam >0$.

\subsection{General regularization scheme}

Recall (see, e.g., Definition 2.2 in \cite{1stbook}) that regularization schemes can be indexed by parametrized functions $g_\lambda : [0,c] \rightarrow \R$,  $\lambda >0.$ The only requirements are that there are positive constants $\gamma_0, \gamma_{- \frac{1}{2}}, \gamma_{-1}$ for which 
\begin{align}\label{A.eq2.1.3}
\begin{split}
\sup_{0 < t \leq c} |1 - t g_\lambda (t)| \leq \gamma_0, \\
\sup_{0 < t \leq c} \sqrt{t} |g_\lambda (t)|\leq \frac{\gamma_{- \frac{1}{2}}}{ \sqrt{\lambda}}, \\
\sup_{0 < t \leq c}  |g_\lambda (t)| < \frac{\gamma_{-1}}{\lambda}.
\end{split}
\end{align}% 
The qualification of the regularization scheme indexed by $g_\lambda$ is the maximal $\nu > 0 $ 
such that for any $\lambda \in (0,c]$ it holds
\begin{align}\label{eq2.1.4}
\sup_{0 < t \leq c} t^\nu | 1 - t g_\lambda (t)  | \leq \gamma_\nu \lambda^\nu, 
\end{align}
where $\gamma_\nu$ does not depend on $\lambda$. Following Definition 2.3 of \cite{1stbook} we also say that qualification $\nu$ covers a non-decreasing function $\varphi : [0,c ] \rightarrow \R$, $\varphi (0) = 0,$ if the function $t \rightarrow \frac{t^\nu}{\varphi (t)}$ is non-decreasing for $t \in (0,c].$ 

Observe that one can use the operator functional calculus to represent IWRLS-approximant (\ref{A.eq2.1.2}) in terms of the function $g_\lambda (t) = (\lambda + t)^{-1}$ indexing Tikhonov regularization. It is easy to check that for $g_\lambda (t) = (\lambda + t)^{-1}$ the requirements (\ref{A.eq2.1.3}) are satisfied with $\gamma_0 = \gamma_{-1} = 1$, $\gamma_{-\frac{1}{2}} = \frac{1}{2}$. Moreover, the qualification $\nu$ of the Tikhonov regularization scheme is equal to $1$, and such a small qualification is the main drawback of this scheme.

Besides, the qualification of the regularization can be increased if one employs the so-called iterated Tikhonov regularization, according to which IWRLS-approach needs to be repeated such that the approximation $f_\mathbf{z}^\lambda = f_{\mathbf{z},l}^\lambda$ obtained in the previous $l$-th step plays the role of an initial guess for the next approximation $f_\mathbf{z}^\lambda = f_{\mathbf{z},l + 1}^\lambda$ constructed as the minimizer of weighted and penalized empirical risk 
\begin{align*}
\mathcal{R}_{\mathbf{z},\lambda, \beta}^{l+1} (f) = \frac{1}{n} \sum_{i = 1}^n \beta_i (f(x_i) - y_i)^2 + \lambda  \norm{f - f_{\mathbf{z},l}^\lambda}_{\mathcal{H}_K}^2, \hspace*{0.3cm} f_{\mathbf{z},0}^\lambda = 0.
\end{align*}
After $\nu$ such iterations we obtain the approximation $f_\mathbf{z}^\lambda = f_{\mathbf{z},\nu}^\lambda$ that can be represented in the form (\ref{A.eq2.1.5}) with 
$$g_\lambda (t) = g_{\lambda,\nu} (t) = \frac{1 - \frac{\lambda^\nu}{(\lambda + t)^\nu}}{t}.$$
The regularization indexed by $g_{\lambda,\nu} (t)$ has the qualification $\nu$ that can be taken as large as desired. Moreover, for $g_\lambda (t) = g_{\lambda,\nu} (t)$ the requirements (\ref{A.eq2.1.3}), (\ref{eq2.1.4}) are satisfied with $\gamma_0 = 1, \gamma_{- \frac{1}{2}} = \nu^{\frac{1}{2}}, \gamma_{-1} = \nu, \gamma_\nu = 1.$

\subsection{General source conditions}
As mentioned in the Introduction, in unsupervised domain adaptation, we intend to approximate a solution of the equation arising from the minimization of the excess risk
	\begin{align*}
	\mathcal{R}_q (f) - \mathcal{R}_q (f_q) = \norm{f - f_q}_{L_{2,\rho_T}}^2
	\end{align*}
Let $J_T: \mathcal{H}_K \hookrightarrow L_{2, \rho_T} $ and $J_S: \mathcal{H}_K \hookrightarrow L_{2, \rho_S} $ be the inclusion operators. Then in ${\mathcal{H}_K} $ the above minimization can be written in terms of the inclusion operator as $\norm{J_T f - f_q}_{L_{2,\rho_T}} \rightarrow min$, and it leads to the infinite-dimensional normal equation
	\begin{align}\label{A.eq2.1.7}
	J_T^* J_T f = J_T^* f_q.
	\end{align}
Due to compactness of the operator $J^* J_T$, its inverse $(J_T^* J_T)^{-1}$ cannot be a bounded operator in ${\mathcal{H}_K} $, and this makes the equation (\ref{A.eq2.1.7}) ill-posed, but since $f_q$ is assumed to be in ${\mathcal{H}_K}  = Range (J_T)$, the Moore-Penrose generalized solution $f^\dagger$ of (\ref{A.eq2.1.7}) coincides in ${\mathcal{H}_K}$  with $f_q$, or $J_T f^\dagger = f_q$ in $L_{2,\rho_T}.$
	
	Of course, the equation (\ref{A.eq2.1.7}) is not accessible because neither $q$ nor $f_q$ are known, but the result \cite{mathe2008} of the regularization theory tells us that there is always a continuous, strictly increasing function $ \varphi : [0,\|J_T^* J_T \|_{\mathcal{H}_K} ] \rightarrow \R$ that obeys $\varphi (0) = 0 $ and allows the representation of $f^\dagger = f_q$ in terms of the so-called source condition:
	\begin{align}\label{A.eq2.1.8}
	f_q = \varphi (J_T^* J_T ) \nu_q, \hspace*{0.3cm} \nu_q \in {\mathcal{H}_K}.
	\end{align}
	The function $\varphi$ above is usually called the index function. Moreover, for every $\epsilon > 0$ one can find such $\varphi$ that (\ref{A.eq2.1.8}) holds true for $\nu_q$ with $$\|\nu_q\|_{\mathcal{H}_K} \leq (1 + \epsilon) \|f_q\|_{\mathcal{H}_K}. $$
	Note that since the operator $J_T^* J_T$ is not accessible, there is a reason to restrict ourselves to consideration of such index functions $\varphi$, which allow us to control perturbations in the operators involved in the definition of source conditions. In the context of supervised learning, a class of such index functions has been discussed in \cite{bauer2007}, and here we follow that study. Namely, we consider the class $\mathcal{F} = \mathcal{F} (0,c)$ of index functions $\varphi : [0,c] \rightarrow \R_+$ allowing splitting $\varphi (t) = \vartheta(t) \psi (t)$ into monotone Lipschitz part $\vartheta, \vartheta(t) = 0,$ with the Lipschitz constant equal to 1, and an operator monotone part $\psi, \psi(0)= 0.$ 
	
	Recall that a function $\psi$ is operator monotone on $[0,c]$ if for any pair of self-adjoint operators $U, V$ with spectra in $[0,c]$ such that $U \leq V$ (i.e. $V - U$ is an non-negative operator) we have $\psi (U) \leq \psi (V)$.
	
	Examples of operator monotone index functions are $\psi (t) = t^\nu,$ $\psi (t) = \log^{-\nu} \left( \frac{1}{t} \right),$ $\psi (t) = \log^{-\nu} \left( \log \frac{1}{t} \right), 0 < \nu  \leq 1,$ while an example of a function $\varphi$ from the above defined class $\mathcal{F}$ is $\varphi(t) = t^r \log^{-\nu} \left( \frac{1}{t} \right), r > 1, 0 < \nu \leq 1,$  since it can be splitted in a Lipschitz part $v(t) = t^r$ and an operator monotone part $\psi (t) = \log^{-\nu} \left( \frac{1}{t} \right).$ Note that source conditions with the above index functions are traditionally considered in the regularization theory. 

 \subsection{Risk bounds under the assumption of knowing the Radon-Nikod\'ym derivative}\label{sec2.3}

Under the assumption that the source condition (\ref{A.eq2.1.8}) holds true, with $\varphi \in \mathcal{F} (0,c)$ and a sufficiently large value of $c$, we consider the approximant $f_\mathbf{z}^\lambda$ as specified in (\ref{A.eq2.1.5}), where the regularization scheme, indexed by $g_\lambda (t)$, has the qualification $\nu$ that covers the function $\varphi (t) \sqrt{t}$. In this context, we establish the risk bounds between the approximant $f_\mathbf{z}^\lambda$ and the target function $f_q$ in RKHS and $L_{2,\rho_S}$ as the following theorem
\begin{theorem}[\cite{gizewski2022regularization}]
\label{A.thm2.1}
Assume that the source condition (\ref{A.eq2.1.8}) is satisfied with $\varphi \in \mathcal{F} (0,c)$ and $c$ is large enough. Consider the approximant $f_\mathbf{z}^\lambda$ given by (\ref{A.eq2.1.5}), where the regularization scheme indexed by $g_\lambda (t)$ has the qualification $\nu$ that covers the function $\varphi (t) \sqrt{t}.$ Consider also the function $\theta (t) = \varphi (t) t$ and choose $\lambda = \lambda_{m,n} = \theta^{-1} (m^{-\frac{1}{2}} + n^{-\frac{1}{2}}).$ Then for sufficiently large $m$ and $n$ with probability at least $1-\delta$ it holds 
\begin{align*}
		\|f_q - f_\mathbf{z}^{\lambda_{m,n} } \|_{L_{2,\rho_T}} &\leq c \log \frac{1}{\delta}\; \varphi (\theta^{-1} (m^{-\frac{1}{2}} + n^{-\frac{1}{2}}))\sqrt{\theta^{-1} (m^{-\frac{1}{2}} + n^{-\frac{1}{2}})}, \\
		\|f_q - f_\mathbf{z}^{\lambda_{m,n} } \|_{\mathcal{H}_K} &\leq c \log \frac{1}{\delta}\; \varphi (\theta^{-1} (m^{-\frac{1}{2}} + n^{-\frac{1}{2}}))
\end{align*}
The values of the coefficients $c$ in the above inequalities do not depend on $\delta, m, n.$
\end{theorem}
To the best of our knowledge, before the study~\cite{gizewski2022regularization}, no error bounds were known even for IWRLS-approach (\ref{A.eq2.1.2}) that corresponds to Tikhonov regularization scheme $g_\lambda (t) = (\lambda + t)^{-1}.$ On the other hand, in the standard supervised learning setting this scheme has been analysed in \cite{smale2007} uniformly for the whole class of RKHS $\mathcal{H}_K$ under the assumption, which in our terms can be written as $\| (J_TJ_T^*)^{-r} f_q \|_{L_{2,\rho_T}} \leq c$ with $r > \frac{1}{2}.$ From Proposition 3.2 of \cite{devito2006} we know that the above assumption can be equivalently written as the source condition (\ref{A.eq2.1.8}) with $\varphi (t) = t^{r - \frac{1}{2}}.$ For this index function our Theorem \ref{A.thm2.1} gives respectively the error bounds of orders $O\left( (m^{-\frac{1}{2}} + n^{-\frac{1}{2}})^{\frac{2r}{2r + 1}}\right)$ and  $O\left( (m^{-\frac{1}{2}} + n^{-\frac{1}{2}})^{\frac{2r-1}{2r + 1}}\right)$ in $L_{2,\rho_T}$ and $\mathcal{H}_K$. For a sufficiently large number $m \geq n$ of unlabeled inputs $x_1',x_2',\ldots,x_m'$ sampled from the target measure $\rho_T$ the above results match the orders of the bounds \cite{smale2007} in the standard supervised learning setting.

The comparison of Theorem \ref{A.thm2.1} with the results \cite{smale2007} (for Tikhonov regularization) and \cite{bauer2007} (for general regularization scheme) allows the conclusion that in the scenario of domain adaptation with covariate shift, one can guarantee the same order of the error as in the standard supervised learning setting, provided that the number of unlabeled target inputs is big enough, and the values of the Randon-Nykodym derivative at that inputs are known.

The latter assumption is seldom satisfied in practice. Therefore, in the next sections, we discuss approximate Randon-Nikod\'ym differentiation and its use in the context of domain adaptation.

\section{Radon-Nikod\'ym differentiation}
\label{sec:radon_Nikod\'ym_differentiation}

Recall that 
our initial assumption in Section~\ref{sec2} has been that the values of the Radon-Nikod\'ym derivative $\beta(x)$ are exactly given.
However, in practice, neither $\rho_S$ nor $\rho_T$ is known.
% Consequently, there arises a need to address the estimation of Radon-Nikod\'ym derivatives.
In this section, we therefore discuss recent results of~\cite{gizewski2022regularization,nguyen2023regularized}, where the goal is to approximate the Radon-Nikod\'ym derivative $\beta = \frac{d\rho_T}{d\rho_S} $ by some function $ \tilde{\beta}$.

We will later use the approximation $ \tilde{\beta}$ within the regularization (\ref{A.eq2.1.5}), where the matrix $B = \mathrm{diag}( \beta (x_1),\beta (x_2),\ldots,\beta (x_n))$ will be substituted by a matrix $\tilde{B} = \mathrm{diag}(\tilde{\beta} (x_1),\tilde{\beta} (x_2),\ldots,\tilde{\beta} (x_n))$ of the corresponding approximate values $\tilde{\beta} (x_i) \approx \beta (x_i), i=1,2,...,n$.
Note, however, that the problem of estimating the Radon-Nikod\'m derivative appears not only in domain adaptation with covariate shift, but also in anomaly detection~\cite{smola2009relative,hido2011statistical}, two-sample testing~\cite{keziou2005test,kanamori2011f}, divergence estimation~\cite{nguyen2007estimating,nguyen2010}, covariate shift adaptation~\cite{shimodaira2000,gizewski2022regularization,dinu2022aggregation}, generative modeling~\cite{mohamed2016learning}, conditional density estimation~\cite{schuster2020}, and  classification from
positive and unlabeled data~\cite{kato2019learning}; cf.~also the monograph~\cite{sugiyama2012book}.

\subsection{Error bounds in RKHS}
In the literature, various RKHS-based approaches are available for a Radon - Nikod\'ym derivative estimation. Here we may refer to \cite{kanamori2012} and to references therein. Conceptually, under the assumption that $\beta \in \hk$, several of the above approaches can be derived from a regularization of an integral equation, which can be written in our terms as 
	\begin{align}\label{A.eq3.1.1}
	J_S^* J_S \beta = J_T^*J_T \mathbf{1} 
	\end{align}
and is ill-pose similar to (\ref{A.eq2.1.7}). Here $\mathbf{1}$ is the constant function that takes the value $1$ everywhere, and almost without loss of generality, we assume that $\mathbf{1} \in {\mathcal{H}_K}$, because otherwise the kernel $K_1 (x, x') = 1 + K(x,x')$ will, for example, be used to generate a suitable RKHS containing all constant functions.

Just as the equation (\ref{A.eq2.1.7}) is inaccessible, so is the equation (\ref{A.eq3.1.1}). But in contrast  to (\ref{A.eq2.1.7}), the reduction of (\ref{A.eq3.1.1}) to a finite-dimensional problem does not require any labels, such as ${\bf y},$ that were necessary for dealing with the normal equation $S_{S_X}^* B S_{S_X}f = S^*_{{\x}} B {\bf y}$. Since in practice, the amount of unlabeled inputs is usually much greater than that of labeled ones, we assume that the sizes $M$ and $N$ of i.i.d. samples $(x_1',x_2',\ldots,x_M')$ and $(x_1,x_2,\ldots,x_N)$ drawn respectively from $\rho_T$ and $\rho_S$ are much larger than $m$ and $n$ appearing in Theorem \ref{A.thm2.1}.
	
Then we consider two sample operators
	\begin{align*}
	&S_{M,T} f = (f(x_1'), f(x_2'),\ldots,f(x_M')) \in \R^M,\\
	&S_{N,S} f = (f(x_1), f(x_2),\ldots,f(x_N)) \in \R^N,
	\end{align*}
	and the finite-dimensional problem
	\begin{align}\label{A.eq3.1.3}
	S_{N,S}^* S_{N,S} \beta = S_{M,T}^* S_{M,T} \mathbf{1},
	\end{align}
which is an empirical version of the equation (\ref{A.eq3.1.1}), where, similar to the above notations the operators $S_{N,S}^* : \R^N \rightarrow {\mathcal{H}_K}, $ $S_{M,T}^*: \R^M \rightarrow {\mathcal{H}_K}$ are given as 
\begin{align*}
	&S_{N,S}^* v(\cdot) = \frac{1}{N} \sum_{i = 1}^N K(\cdot, x_i) v_i,\hspace*{0.5cm} u = (v_1,v_2,\ldots,v_N) \in \R^N,\\
	&S_{M,T}^* u(\cdot) = \frac{1}{M} \sum_{j = 1}^M K(\cdot, x_j') u_j, \hspace*{0.5cm} u = (u_1,u_2,\ldots,u_M) \in \R^M.
\end{align*}
A regularization of equations (\ref{A.eq3.1.1}), (\ref{A.eq3.1.3}) may serve as a starting point for several approaches of estimating the Radon - Nikod\'ym derivative $\beta$. For example, as it has been observed in \cite{kanamori2012}, the known kernel mean matching (KMM) method \cite{huang2006} can be viewed as the regularization of (\ref{A.eq3.1.1}), (\ref{A.eq3.1.3}) by the method of quasi (least-squares) solutions, originally proposed by Valentin Ivanov (1963) and also known as Ivanov regularization (see, e.g., \cite{oneto2016} and \cite{page2019} for its use in the context of learning).  In KMM
an Ivanov-type regularization is applied to the empirical version (\ref{A.eq3.1.3}) and  leads to a quadratic problem.  At the same time, the kernelized unconstrained least-squares importance fitting (KuLSIF) proposed in \cite{kanamori2012} allows an analytic-form solution and can be reduced to solving a linear problem with respect to corresponding variables.
	
From Theorem 1 of \cite{kanamori2012} it follows that in KuLSIF the approximation $\tilde{\beta} $ of the Radon - Nikod\'ym derivative $\beta = \frac{d\rho_T}{d\rho_S}$ is in fact constructed by application of the Tikhonov regularization scheme to the empirical version (\ref{A.eq3.1.3}) of the equation (\ref{A.eq3.1.1}), that is in KuLSIF we have 
	\begin{align} \label{A.eq3.1.4}
	\tilde{\beta} = \beta_{M,N}^\lambda = g_\lambda (S_{N,S}^* S_{N,S}) S_{M,T}^*S_{M,T} \mathbf{1},
	\end{align}
	where $g_\lambda (t) = (\lambda + t)^{-1}.$

Though there are several studies devoted to KMM and KuLSIF, to the best of our knowledge there has been no study of pointwise approximation error $\beta (x) - \tilde{\beta} (x)$, which is of interest in analysis of regularized domain adaptation methods, such as IWRLS. For example, in \cite{kanamori2012} and \cite{que2013} (see Type I setting there) the statistical consistency and accuracy of KuLSIF have been analysed in the space $L_{2,\rho_S}$, where pointwise evaluations are undefined. We can also mention the study \cite{schuster2020}, where KuLSIF represented as (\ref{A.eq3.1.4}) with $g_\lambda (t) = (\lambda + t)^{-1}$ was discussed in a RKHS, but only convergence of $\tilde{\beta}$ to $\beta$ was proved, without quantifying its rate.
	
At the same time, using the concept of source conditions naturally appearing because of equation (\ref{A.eq3.1.1}), we can obtain the following statement

\begin{theorem}[\cite{gizewski2022regularization}]
\label{thm3.1}
		Assume that $\beta = \frac{d\rho_T}{d\rho_S}$ meets source condition $\beta = \phi (J_S^* J_S) \nu_\beta$, where $\phi \in \mathcal{F} (0,c),$ and $c$ is large enough. Consider the approximant $\beta_{M,N}^\lambda$ given by (\ref{A.eq3.1.4}), where the regularization scheme indexed by $g_\lambda (t)$ has the qualification $\nu$ that covers the index function $\phi (t)$. Let $\lambda = \lambda_{M,N} = \theta_\phi^{-1} (M^{-\frac{1}{2}} + N^{-\frac{1}{2}}),$ where $\theta_\phi (t) = \phi (t) t.$ Then for sufficiently large $M$ an $N$ with probability at least $1 - \delta$ it holds
		\begin{align}\label{1st_bound_beta}
		\norm{\beta - \beta_{M,N}^{\lambda_{M,N}} }_{\mathcal{H}_K} \leq c \left( \log \frac{1}{\delta} \right) \phi \left( \theta_\phi^{-1} (M^{-\frac{1}{2}} + N^{-\frac{1}{2}}) \right).
		\end{align}
\end{theorem}

\begin{remark}\label{rem_new2}
	As we already mentioned, for $ g_\lambda (t) = (\lambda + t)^{-1} $ the convergence in probability of the approximation (\ref{A.eq3.1.4}) to $ \beta $ in RKHS has been proven in \cite{schuster2020}. Such convergence has been established in \cite{schuster2020} on the base of an error estimation (see Supplementary material A in \cite{schuster2020}), which in our terms can be written as follows:
	\begin{align}\label{new_eq2}
	\norm{\beta - \beta_{M,N}^\lambda}_{\mathcal{H}_K} \leq \norm{\beta - \beta^\lambda}_{\mathcal{H}_K} + c \left( \frac{N^{-a}}{\lambda^2} + \frac{M^{-b}}{\lambda}\right) \left( \log \frac{1}{\delta}\right),
	\end{align}
	where $ \beta^\lambda = (\lambda I + J_S^*J_S)^{-1} J_T^*J_T \mathbf{1}, $ and $ 0 < a < \frac{1}{2}, 0<b< \frac{1}{2}.$
	Note that in the regularization theory, the quantities $ \norm{\beta - \beta^\lambda} $ are sometimes called the profile functions, and Corollary 2 of \cite{mathe2008} estimates them in terms of the source condition $ \beta = \phi(J_S^*J_S)\nu_\beta $ as follows
	\begin{align}\label{new_eq3}
	\norm{\beta - \beta^\lambda}_{\mathcal{H}_K} \leq c \phi(\lambda).
	\end{align}
Furthermore, the bound presented in Theorem \ref{thm3.1} can be expressed in the following form
\begin{align}\label{new_eq1}
\norm{\beta - \beta_{M,N}^\lambda}_{\mathcal{H}_K} \leq c \left( \phi (\lambda) + \frac{M^{-\frac{1}{2}} + N^{-\frac{1}{2}}}{\lambda} \right) \log \left( \frac{1}{\delta}\right) .
\end{align}
Comparing (\ref{new_eq2}) - (\ref{new_eq1}) and keeping in mind that $ \frac{M^{-\frac{1}{2}} + N^{-\frac{1}{2}} }{\lambda} $ is smaller in the sense of the order than $ \left( \frac{N^{-a}}{\lambda^2} + \frac{M^{-b}}{\lambda}  \right) $, one can conclude that the error bound (\ref{new_eq1}) obtained by our argument generalized and refines the results of \cite{schuster2020}.
\end{remark}
\begin{remark}\label{rem_new3}
	Recall that for our purpose we restrict ourselves to the estimation of accuracy of (\ref{A.eq3.1.4}) in RKHS. At the same time, there are studies, where the accuracy of (\ref{A.eq3.1.4}) has been analysed in the space $ L_{2,\rho_S} $. For example, in \cite{kanamori2012} $ L_{2,\rho_S} $ - convergence rate of KuLSIF estimator, which corresponds to (\ref{A.eq3.1.4}) with $ g_\lambda (t) = (\lambda + t)^{-1}, $ has been established in terms of the order $ \gamma $ of the so-called bracketing entropy of the underlying space $ \mathcal{H}_K $. For a given space $ \mathcal{H}_K $, the $ \gamma $- value is fixed within the interval $ (0,2) $, i.e., $ 0<\gamma<2, $ and Theorem 2 of \cite{kanamori2012} tells that if $ \beta \in \mathcal{H}_K,  $ then for arbitrary $ \epsilon $ satisfying $ 1 - \frac{2}{(2+\gamma)} < \epsilon < 1,$ and $ \lambda_{M,N,\epsilon}^{-1} = O \left( (N \wedge M)^{1-\epsilon}\right) $ with high probability it holds 
	\begin{align}\label{new_eq4}
	\norm{\beta - \beta_{M,N}^{\lambda_{M,N,\epsilon}}}_{L_{2,\rho_S}} = O \left( (N \wedge M)^{- \frac{(1-\epsilon)}{2}}\right), 
	\end{align}
	where $ N \wedge M = \min \{N,M\} $. Note that the rate established by (\ref{new_eq4}) cannot be better than $ O\left( (N \wedge M)^{- \frac{1}{2 + \gamma}} \right) $, and it does not take into account additional smoothness that any particular element $ \beta \in \mathcal{H}_K $ has in the underlying space $\mathcal{H}_K $. Such additional smoothness can be caught in the form of a source condition because as we already know from \cite{mathe2008}, there is always an index function $ \phi $ such that $ \beta = \phi (J_S^*J_S) \nu_\beta. $ Then assuming that $ \phi (t) \sqrt{t} $ is covered by the qualification of the regularization used in (\ref{A.eq3.1.4}), %and applying second part of Lemma \ref{lem2.2} in the same way as in the proof of Theorem \ref{thm3.1}, 
 and applying almost the same argument as in Theorem \ref{thm3.1} we can obtain the bound 
	\begin{align}\label{new_eq5}
	\norm{\beta - \beta_{M,N}^{\lambda_{M,N}}}_{L_{2,\rho_S}} \leq c \left( \left( \log \frac{1}{\delta}  \right) \phi \left( \theta_\phi^{-1} (M^{-\frac{1}{2}} + N^{-\frac{1}{2}})\right) \right) \sqrt{\theta_\phi^{-1} (M^{-\frac{1}{2}} + N^{-\frac{1}{2}})}.
	\end{align}
	To simplify a comparison of (\ref{new_eq5}) with the best possible rate $ O \left( (N \wedge M)^{- \frac{1}{2 + \gamma}} \right) $ of (\ref{new_eq4}) we consider the same index functions $ \phi (t) = t^{r - \frac{1}{2}} $ as in Section \ref{sec2.3}. Then (\ref{new_eq5}) gives the rate of order $ O \left( (M^{-\frac{1}{2}} + N^{-\frac{1}{2}})^{\frac{2r}{2r+1}} \right) $, and for $ r > \frac{1}{\gamma} $ this rate is better than the one given by (\ref{new_eq4}), because 
	$$(N \wedge M)^{- \frac{1}{2 + \gamma}}  > (M^{-\frac{1}{2}} + N^{-\frac{1}{2}})^{\frac{2}{2+\gamma}}  > (M^{-\frac{1}{2}} + N^{-\frac{1}{2}})^{\frac{2r}{2r+1}}. $$
	This is one more example of how the study~\cite{gizewski2022regularization} generalizes, specifies and refines previously known results.
\end{remark}

In~\cite{nguyen2023regularized}, we highlight that the convergence of algorithms for Radon-Nikod\'ym differentiation is impacted by both, the smoothness of the function being approximated and the capacity of the approximating space. However, the result in Theorem \ref{thm3.1} only takes into consideration the smoothness of the approximated derivative $\beta$. Therefore, we follow \cite{pauwels2018} and employ the concept of the so-called regularized Christoffel function that allows direct incorporation of the regularization parameter $\lam$ into the definition of a capacity measure.
Consider the function 
\begin{align}\label{def:Chris_func}
	C_\lam (x) = \innerpro{K(\cdot,x), (\lambda I +J_{S}^* J_{S} )^{-1} K(\cdot,x) }_{\mathcal{H}_K} = \norm{(\lambda I + J_S^* J_S)^{-\frac{1}{2}} K(\cdot,x) }^2_{\mathcal{H}_K}
\end{align}
Note that in \cite{pauwels2018} the reciprocal of $C_\lam (x)$, \ie $\frac{1}{C_\lam (x)}$, was called the regularized Christoffel function, but for the sake of simplicity, we will keep the same name also for (\ref{def:Chris_func}). Note also that in the context of supervised learning, where usually only one probability measure, say $p$, is involved, the expected value
\[
\mathcal{N} (\lam) = \int_\bX C_\lam (x) d p(x)
\]
of $C_\lam (x)$, called the effective dimension, has been proven to be useful \cite{caponnetto2007}. This function is also frequently used as a capacity measure of $\hk$.

At the same time, if more than one measure appears in the supervised learning context, as is, for example, the case in the analysis of Nystr{\"o}m subsampling \cite{rudi2015, shuai2019Analysis}, then the $C$-norm of the regularized Christoffel function 
\begin{align}\label{def:N_inf}
	\mathcal{N}_\infty (\lambda) := \sup_{x \in \mathbf{X}} C_\lambda (x) 
\end{align}
is used in parallel with the effective dimension $\mathcal{N} (\lambda)$. This gives a hint that $\mathcal{N}_\infty (\lambda)$ could also be a suitable capacity measure for analysing the accuracy of Radon-Nikod\'ym numerical differentiation, since there more than one measure is also involved.

To estimate the regularized Christoffel functions we slightly generalize a source condition for kernel sections $K(\cdot,x)$ that has been used in various contexts in \cite{shuai2019Analysis} and \cite{devito2014}.
\begin{assumption} \label{assum:source_cond_kernel}
	(Source condition for kernel) There is an operator concave index function $\xi : [0, \norm{J^*_S J_S}] \to [0,\infty)$ such that  $\xi^2$ is covered by the qualification $\nu=1$, and for all $x \in \bX $,
	\begin{align*}
	    K(\cdot, x) = \xi (J^*_S J_S) v_x, \hspace*{0.3cm} \norm{v_x}_\hk \leq c, 
	\end{align*}
	where $c$ does not depend on $x$.	
\end{assumption}
We mention the following consequence of Assumption \ref{assum:source_cond_kernel}.
\begin{lemma}
\label{lem:N_infty_bound}
	Under Assumption \ref{assum:source_cond_kernel},
	\begin{align*}
	    \mathcal{N}_\infty (\lambda) \leq c \frac{\xi^2(\lambda)}{\lambda}. 
	\end{align*}
\end{lemma}
By taking into account the smoothness properties of the Radon-Nikod\'ym derivative $\beta$ and the capacity of $\hk$ expressed in terms of the regularized Christoffel functions, in~\cite{nguyen2023regularized}, we establish a novel bound that relates $\beta_{M, N}^\lambda$ to $\beta$ in RKHS.

\begin{theorem}[\cite{nguyen2023regularized}]
\label{thm4.5}
	Let $K$ satisfies Assumption \ref{assum:source_cond_kernel}, and $\lam \geq \lam^*$ with $\lam_*$ satisfying $\mathcal{N}(\lam_*)/\lam_* = N$. If $\beta = \frac{d\rho_T}{d\rho_S}$ meets source condition (\ref{A.eq2.1.8}) $\beta = \phi (J_S^* J_S) \nu_\beta$, where $\phi = \vartheta \psi \in \mathcal{F} (0,c)$ with large enough $c$, and the qualification $g_\lam$ covers $\vartheta (t) t^\frac32$, then with probability at least $1 - \delta$, it holds
	\begin{align*}
		\norm{\beta - \beta^\lam_{M,N}}_\hk \leq c \left( \phi(\lam) +  (M^{-\frac12} + N^{-\frac12}) \frac{\xi (\lam)}{\lam} \right) \left( \log \frac{2}{\delta}\right)^2.
	\end{align*}
	Consider $\theta_{\phi,\xi} (t) = \frac{\phi(t) t}{\xi(t)}$ and $\lam = \lambda_{M,N} = \theta_{\phi,\xi}^{-1} (M^{-\frac12} + N^{-\frac12})$, then
	$$\norm{\beta - \beta^\lam_{M,N}}_\hk \leq c \phi\left(\theta_{\phi,\xi}^{-1} (M^{-\frac12} + N^{-\frac12})\right) \log^2 \frac1\delta.$$
\end{theorem}

\begin{remark}\label{rem_sec4}
    In order to compare the bounds presented in Theorem \ref{thm3.1} and Theorem \ref{thm4.5}, let us consider the case when $\beta$ meets the source condition (\ref{A.eq2.1.8}) with $\phi (t) = t^\eta$. 
    %$\eta\geq\frac12$. 
     %If $\lam = \lam_{m,n} = \theta_\varphi^{-1} (m^{-\frac12} + n^{-\frac12}) $, 
     In this case the bound (\ref{1st_bound_beta}) can be reduced to 
	\begin{align}\label{rate_in_acha}
		\norm{\beta - \beta^\lam_{M,N}}_\hk  = O \left((M^{-\frac12} + N^{-\frac12})^\frac{\eta }{\eta +1}\right).
	\end{align}
	It is noteworthy that the error bound established in Theorem \ref{thm3.1} does not take into consideration the capacity of $\hk$. Such an additional factor can be accounted for in terms of Assumption \ref{assum:source_cond_kernel}. Assume that $K$ satisfies Assumption 3.2 with $\xi (t) = t^\varsigma, 0< \varsigma  \leq \frac12$, then for $\lam = \lam_{M,N} = \theta_{\phi,\xi}^{-1} (M^{-\frac12} + N^{-\frac12})$, the bound in Theorem \ref{thm4.5} gives  
	\[
	\norm{\beta - \beta^\lam_{M,N}}_\hk = O \left((M^{-\frac12} + N^{-\frac12})^\frac{\eta }{\eta +1 - \varsigma } \right),
	\]
%	It is easy to see that for $\varsigma > 0$ this 
 that is better than the order of accuracy given by (\ref{rate_in_acha}). Then one can conclude that the bound in Theorem \ref{thm4.5} obtained by our argument generalizes, specifies, and refines the results in Theorem \ref{thm3.1}. 
\end{remark}

\subsection{Error bounds for the pointwise evaluation}
In the following, we discuss the error between point values of $\beta (x)$ and $\beta^\lam_{M,N} (x) $ for any $x \in \bX$. In view of the reproducing property of $K$ we have 
\begin{align*}
	|\beta (x) - \beta^\lam_{M,N} (x) | = \big|  \innerpro{K_{x},\beta - \beta^\lam_{M,N} }_\hk \big| 
	& = \big|  \innerpro{K(\cdot , x),\beta - \beta^\lam_{M,N} }_\hk \big|\nonumber\\
	&= \big| \innerpro{\xi(J^*_S J_S)v_x, \beta - \beta^\lam_{M,N}}_\hk \big| \nonumber\\
	&\leq c \norm{\xi (J^*_S J_S) (\beta - \beta^\lam_{M,N})  }_\hk. %\label{eq:pw_beta}
\end{align*}
%Similarly, we obtain 
%\begin{align*}
%	|\beta (x) - \beta^\lam (x) | &\leq c \norm{\xi (J^*_S J_S) (\beta - \beta^\lam)}_\hk, \\
%	|\beta^\lam (x) - \beta^\lam_{M,N} (x) | &\leq c \norm{\xi (J^*_S J_S) (\beta^\lam - \beta^\lam_{M,N}) }_\hk,
%\end{align*}
%that allows for the following decomposition of the error bound
%\begin{align*}%\label{eq:pw_decom_bound}
%		|\beta (x) - \beta^\lam_{M,N} (x) | \leq c \left(\norm{\xi (J^*_S J_S) (\beta - \beta^\lam)}_\hk + \norm{\xi (J^*_S J_S)%(\beta^\lam - \beta^\lam_{M,N}) }_\hk \right).
%\end{align*}
%Subsequently, we establish the risk bounds in Paper C as the following theorem.
Thus, the error between point values can be bounded by the approximation error in a weighted norm that is weaker than the one of the underlying space. Therefore, pointwise error estimates can be smaller than the ones guaranteed by Theorem \ref{thm4.5}. This observation is quantified by the following theorem in~\cite{nguyen2023regularized}.
\begin{theorem}[\cite{nguyen2023regularized}]
\label{thrm5.3}
		Under the assumption of Theorem \ref{thm4.5}, for $\lambda > \lam_*$ with probability at least $1 - \delta$, for all $x \in \bX$, we have 
			\begin{align*}
			|\beta (x) - \beta^\lam_{M,N} (x) |  \leq c \xi (\lam) \left( \phi(\lam) +  (M^{-\frac12} + N^{-\frac12}) \frac{\xi (\lam)}{\lam} \right) \left( \log \frac{2}{\delta}\right)^2,
		\end{align*}
	and for $\lam = \lambda_{M,N} = \theta_{\phi,\xi}^{-1} (M^{-\frac12} + N^{-\frac12}),$
	\[ |\beta (x) - \beta^\lam_{M,N} (x) |  \leq c \xi (\lambda_{M,N}) \phi(\lambda_{M,N}) \log^2 \frac{1}{\delta}.\]
\end{theorem}

\begin{remark}\label{rem5.4}
%	As has been emphasized in Remark \ref{rem_sec4}, the accuracy of (\ref{eq:beta_g_lam}) has been analyzed in RKHS with the best possible rate $O \left((M^{-\frac12} + N^{-\frac12})^\frac{\eta }{\eta +1 - \varsigma } \right)$. 
 
 Let us consider the same index functions $\varphi (t) = t^\eta$ and $\xi (t)= t^\varsigma$ as in Remark \ref{rem_sec4}, where the accuracy of order $O \left((M^{-\frac12} + N^{-\frac12})^\frac{\eta }{\eta +1 - \varsigma } \right)$ has been derived for (\ref{A.eq3.1.4}). Under the same assumptions, Theorem \ref{thrm5.3} guarantees the accuracy of order $O \left((M^{-\frac12} + N^{-\frac12})^\frac{\eta +\varsigma}{\eta +1 - \varsigma } \right)$.  This illustrates that the reconstruction of the Radon-Nikod\'ym derivative at any particular point can be done with much higher accuracy than its reconstruction as an element of RKHS.
But it should be stressed that the above high order of accuracy is guaranteed when the qualification $\nu$ of the used regularization scheme is higher than that of the Tikhonov–Lavrentiev regularization employed in KuLSIF.
\end{remark}

\section{Embedded regularization}
\label{sec:embedded_regularization}
In this section, we derive a novel error bound based on the combination of the results discussed in last two sections.
More precisely, in Section~\ref{sec2}, we have established  error bounds between the approximant $f_\bz^\lambda$ and the target function $f_q$ in the context of domain adaptation, assuming knowledge of the exact value of the Radon-Nikod\'ym derivative $\beta (x)$.
However, this assumption is rarely satisfied in practice.
Consequently, we embed the regularized Radon-Nikod\'ym numerical differentiation 
considered in the previous section 
into the general regularization scheme for unsupervised domain adaptation such that no exact values of $\beta (x)$ are required. 

This is done by substituting the matrix $B$ in (\ref{A.eq2.1.5}) by the matrix $$B_{M,N} = \mathrm{diag}(\beta_{M,N}^{\lambda_{M,N}} (x_1),\beta_{M,N}^{\lambda_{M,N}} (x_2),\ldots,\beta_{M,N}^{\lambda_{M,N}} (x_n) ).$$
As the result, we obtain an embedded regularization, which produces $f_{\bz,M,N}^{\lambda}$ instead of $f_\bz^\lambda$.
With the exact same arguments as used in Section~3.2 of~\cite{gizewski2022regularization}, in the theorem below we obtain the error bounds for such embedded regularization.
The only difference is that, instead of utilizing the bound from Theorem~\ref{thm3.1} as done in~\cite[Section~3.2]{gizewski2022regularization}, we employ the error bounds derived in Theorem~\ref{thrm5.3}. 
\begin{theorem}\label{new_thm4.1}
		Let assumptions and conditions of Theorems \ref{A.thm2.1} and \ref{thrm5.3} be satisfied. Then with probability at least $1 - \delta$ for 
		$$\lambda_\delta = \theta^{-1} \left( m^{-\frac{1}{2}} + n^{-\frac{1}{2}} + \xi (\theta_{\phi,\xi}^{-1} (M^{-\frac{1}{2}}  + N^{-\frac{1}{2}}) ) \phi (\theta_{\phi,\xi}^{-1} (M^{-\frac{1}{2}}  + N^{-\frac{1}{2}} ) ) \right),$$
		it holds
    \begin{align*}
		&\norm{f_q - f_{\bz,M,N}^{\lambda_\delta}    }_{\mathcal{H}_K} \leq c \left( \log^{\frac{3}{2} } \frac{1}{\delta} \right) \varphi \left(\lambda_\delta \right),\\
		&\norm{ f_q - f_{\bz,M,N}^{\lambda_\delta} }_{L_{2,\rho_T}} \leq c \left( \log^{\frac{3}{2} } \frac{1}{\delta} \right) \varphi \left(\lambda_\delta  \right) \sqrt{\lambda_\delta}.
		\end{align*} 
	\end{theorem}

\begin{remark}\label{rem4.1}
    As has been emphasized in Remark 4 of~\cite{gizewski2022regularization}, the main message of Theorem 3 in~\cite{gizewski2022regularization} is that in unsupervised domain adaptation the error bounds of the same order as in the standard supervised learning may potentially be guaranteed provided that there are big enough amounts of unlabeled data sampled from both target and source domains. To estimate how big these amounts have to be, let us consider $\beta = \frac{d \rho_T}{d\rho_S}$ meeting the so-called Hölder type source conditions $\beta = \phi (J_S^*J_S) \nu_\beta$ with $\phi (t) = t^a$, where the Hölder exponent $a$ can be arbitrary small but positive to guarantee the inclusion of $\beta$ in $\hk$. 
    %In the interest of taking 
    To take into account the capacity of $\hk$
    %, which can be accounted in terms of Assumption \ref{assum:source_cond_kernel}, 
    we assume that $K$ satisfies Assumption 3.2 with $\xi (t) = t^b, 0<b \leq \frac{1}{2}$, then $\xi (\theta_{\phi,\xi}^{-1} (M^{-\frac{1}{2}}  + N^{-\frac{1}{2}}) ) \phi (\theta_{\phi,\xi}^{-1} (M^{-\frac{1}{2}}  + N^{-\frac{1}{2}} ) ) = (M^{-\frac{1}{2}}  + N^{-\frac{1}{2}} )^{\frac{a+b}{a-b+1}} $. Now it is interesting to observe that if  $b=\frac{1}{2}$, then independently of the Hölder exponent $a$, the error bounds guaranteed by our novel Theorem \ref{new_thm4.1} will be of the same order as the ones in Theorem \ref{A.thm2.1}, provided $M$ and $N$ are respectively of order $m$ and $n$. This is essential improvement compared to Remark 4 of~\cite{gizewski2022regularization}, where it was required that an amount of unlabeled data should be at least as big as the squared amount of labeled ones to potentially allow an accuracy of the same order as in the standard supervised learning.
    %\textcolor{red}{Moreover, these error bounds are independent of the smoothness of $\beta$ with $ b = \frac12 $ and $\beta \in \hk$.} 
    %Therefore, one can conclude that the bound in Theorem \ref{new_thm4.1} obtained by our argument generalizes, specifies, and refines the results in Theorem 3 of Paper A. 
\end{remark}

\section{Parameter choice}
\label{sec:parameter_choice}

To complete our analysis of IWRLS, in the following, we discuss the strategies presented in~\cite{gizewski2022regularization,dinu2022aggregation} for choosing regularization parameters.
In particular, we obtain a novel error bound for these methods that is based on Theorem~\ref{new_thm4.1}.

The regularization parameters $\lambda_{m,n}, \lambda_{M,N}, \lambda_\delta$ used by Theorems~\ref{A.thm2.1}, \ref{thrm5.3}, and~\ref{new_thm4.1} crucially rely on the knowledge of the index functions $\varphi, \phi $ and $\xi$, where $\varphi, \phi $ describes the smoothness of $f_q, \beta = \frac{d\rho_T}{d\rho_S}$ in terms of the corresponding source conditions, and $\xi$ describes the capacity of $\hk$. Since such smoothness and capacity of approximated space are usually unknown, one faces the issue of choosing the values of the regularization parameters used  
for constructing the approximations $f_{\mathbf{z},M,N}^{\lambda}$.
The idea in~\cite{gizewski2022regularization,dinu2022aggregation} is to construct a linear combination 
	\begin{align}\label{eq4.1.2}
	f_\mathbf{z} = \sum_{k = 1}^l c_k f_{\mathbf{z},M,N}^{\lambda_k}
	\end{align}
of approximants corresponding to all tried values of the regularization parameters used in the embedded regularization  $f_{\mathbf{z},M,N}^{\lambda}$. For the sake of simplicity we still label that approximants by a sequence of $\{\lambda_k\}_{k=1}^l$.
%given by (\ref{A.eq2.1.5}) for all tried values of the regularization parameter $\lambda = \lambda_1, \lambda_2, \ldots,\lambda_l.$ 

It is clear that the best $L_{2,\rho_T}-$space approximation of the target regression function $f_q$ by the above linear combinations $f_\mathbf{z}$ corresponds to the vector ${{\bf c}} = (c_1,c_2,\ldots,c_l)$ of ideal coefficients in (\ref{eq4.1.2}) that solves the linear system $G{{\bf c}} = {{\bf g}}$ with the Gram matrix $G = \left(  \innerpro{ f_{\mathbf{z},M,N}^{\lambda_k}, f_{\mathbf{z},M,N}^{\lambda_u}}_{L_{2,\rho_T}} \right)_{k,u=1}^l$ and the right-hand side vector ${{\bf g}} = \left( \innerpro{f_q , f_{\mathbf{z},M,N}^{\lambda_k} }_{L_{2,\rho_T}} \right)_{k=1}^l.$ But, of course, neither Gram matrix $G$ nor the vector ${{\bf g}}$ is accessible, because there is no access to the target measure $\rho_T.$
	
To overcome this obstacle we first observe that the norms $\norm{f_{\mathbf{z},M,N}^{\lambda_k}  }_{\mathcal{H}_K}  $ are under our control, such that we can put a threshold $\gamma_l > 0 $ and consider only such approximants
%$\lambda_k, k = 1,2, \ldots, l,$ 
for which $\norm{f_{\mathbf{z},M,N}^{\lambda_k}  }_{\mathcal{H}_K} \leq \gamma_l. $

In practice the number $l$ of the elements in the set $\{ f_{\mathbf{z},M,N}^{\lambda_k} \}_{k=1}^l$ can be assumed to be negligible compared to the cardinalities $m,n,M,N$ of the available data samples (usually not more than 10 - 15 approximants are computed for different values of the regularization parameters). Therefore, $l$-dependent coefficients do not affect the orders $O\left(  m^{-\frac{1}{2}} + n^{-\frac{1}{2}} + \xi (\theta_{\phi,\xi}^{-1} (M^{-\frac{1}{2}}  + N^{-\frac{1}{2}}) ) \phi (\theta_{\phi,\xi}^{-1} (M^{-\frac{1}{2}}  + N^{-\frac{1}{2}} ) )\right)$. Then the inaccessible Gram matrix $G$ and the vector ${{\bf g}}$ can be approximated by respectively 
    \begin{align*}
	\tilde{G} = \left( \frac{1}{m} \sum_{j = 1}^m f_{\mathbf{z},M,N}^{\lambda_k} (x_j') f_{\mathbf{z},M,N}^{\lambda_u} (x_j') \right)_{k,u = 1}^l,\hspace*{0.1cm}
	\tilde{g} = \left( \frac{1}{n} \sum_{i = 1}^n \tilde{\beta}_{M,N} (x_i) y_i f_{\mathbf{z},M,N}^{\lambda_k} (x_i) \right)_{k=1}^l,
	\end{align*}
which can be effectively computed from data samples. Under the assumption that $\tilde{G}^{-1}$ exists, we can approximate function $f_\mathbf{z}$ by $\tilde{f}_z = \sum_{k = 1}^l \tilde{c}_k f_{\mathbf{z},M,N}^{\lambda_k},$ where $\tilde{c} = (\tilde{c}_1,\tilde{c}_2,\ldots,\tilde{c}_l) = \tilde{G}^{-1} \tilde{g}$.

With the same arguments as used in Section~4 of~\cite{gizewski2022regularization}, we establish a novel error bound between $\tilde{f}_z$ and $f_\mathbf{z}$ in space $L_{2,\rho_T}$ as the following theorem.
The only difference to~\cite[Section~4]{gizewski2022regularization} is that we use our improved bounds of Theorem \ref{new_thm4.1} for their Eq.~(24).
\begin{theorem}\label{thm4.1}
	Assume that the approximants are such
% for $\lambda_1,\lambda_2,\ldots,\lambda_l$ 
 that $\norm{f_{\mathbf{z},M,N}^{\lambda_k}  }_{\mathcal{H}_K} \leq \gamma_l,$
% ;  k = 1,2,\ldots,l, $ 
 and the conditions of Theorems \ref{A.thm2.1} and \ref{thrm5.3} hold. Consider $\tilde{f}_z = \sum_{k = 1}^l \tilde{c}_k f_{\mathbf{z},M,N}^{\lambda_k},$ where $\tilde{c} = (\tilde{c}_1,\tilde{c}_2,\ldots,\tilde{c}_l) = \tilde{G}^{-1} \tilde{g}$.
 Then with probability $1 -\delta$ it holds that%
\begin{align}
\label{eq4.1.8}
\begin{split}
&\norm{f_q - \tilde{f}_z}_{L_{2,\rho_T}} \leq \min_{c_k} \norm{f_q - \sum_{k = 1}^l c_k f_{\mathbf{z},M,N}^{\lambda_k}}_{L_{2,\rho_T}}\\
&~~~~+ C\cdot\log\!\left( \frac1\delta \right) \left(  m^{-\frac{1}{2}} + n^{-\frac{1}{2}} + \xi (\theta_{\phi,\xi}^{-1} (M^{-\frac{1}{2}}  + N^{-\frac{1}{2}}) ) \phi (\theta_{\phi,\xi}^{-1} (M^{-\frac{1}{2}}  + N^{-\frac{1}{2}} ) )\right),
\end{split}
\end{align}%
for a generic constant $C>0$ maybe depending on $l$ but not depending on $m,n,M,N.$
\end{theorem}
Assume that the sequence $\lambda_1,\lambda_2,\ldots,\lambda_l$ 
%of the tried values of the regularization parameter $\lambda$ 
is so tight that one of the values, say $\lambda = \lambda_\mu$, is so close to the value $\lambda_\delta$ indicated in Theorem \ref{new_thm4.1}, and the corresponding approximant $f_{\mathbf{z},M,N}^{\lambda_\mu}$ provides an accuracy of the order guaranteed by that theorem. Then under conditions of Theorem \ref{thm4.1} the aggregate approximation $\tilde{f}_z$ also guarantees an accuracy of the same order but does not require any knowledge of the index functions $\varphi, \phi$ describing the smoothness of $f_q$ and $\frac{d\rho_T}{d\rho_S}$, and the index function $\xi$ describing the capacity of $\hk$. This follows from the fact that the second term of the right-hand side of (\ref{eq4.1.8}) is negligible compared to the error bounds given by Theorem \ref{new_thm4.1}, and from the obvious inequality
\begin{align*}
\min_{c_k} \norm{f_q - \sum_{k = 1}^l c_k f_{\mathbf{z},M,N}^{\lambda_k} }_{L_{2,\rho_T}} \leq \norm{f_q - f_{\mathbf{z},M,N}^{\lambda_\mu} }_{L_{2,\rho_T}}.
\end{align*}

\section{Conclusion and outlook}

In this work, we discussed recent error bounds for IWRLS in the setting of unsupervised domain adaptation with covariate shift.
Such error bounds require a combined study of weighted kernel ridge regression with methods for estimating Radon-Nikod\'ym derivatives, as the regression is applied on the output of the estimation procedures.
As a novel result, we obtained, as combination of known results, novel error bounds.
It turned out that a weak source condition on the reproducing kernel allows IWRLS to achieve the same order of accuracy as kernel ridge regression in standard supervised learning, with a much smaller number of samples than anticipated before.

\begin{acknowledgement}
The research reported in this paper has been funded by the Federal Ministry for Climate Action, Environment, Energy, Mobility, Innovation and Technology (BMK), the Federal Ministry for Digital and Economic Affairs (BMDW), and the Province of Upper Austria in the frame of the COMET--Competence Centers for Excellent Technologies Programme and the COMET Module S3AI managed by the Austrian Research Promotion Agency FFG.
\end{acknowledgement}

% \setmaxbibnames{999}
\bibliographystyle{plain}
\bibliography{references.bib}

\begin{thebibliography}{10}

\bibitem{bauer2007}
Frank Bauer, Sergei Pereverzev, and Lorenzo Rosasco.
\newblock On regularization algorithms in learning theory.
\newblock {\em Journal of complexity}, 23(1):52--72, 2007.

\bibitem{caponnetto2007}
A.~Caponnetto and E.~De~Vito.
\newblock Optimal rates for the regularized least-squares algorithm.
\newblock {\em Foundations of Computational Mathematics}, 7(3):331--368, 2007.

\bibitem{Cucker2001OnTM}
Felipe Cucker and Steve Smale.
\newblock On the mathematical foundations of learning.
\newblock {\em Bulletin of the American Mathematical Society}, 39:1--49, 2001.

\bibitem{devito2006}
Ernesto De~Vito, Lorenzo Rosasco, and Andrea Caponnetto.
\newblock Discretization error analysis for {T}ikhonov regularization in
  learning theory.
\newblock {\em Analysis and Applications}, 4:81--99, 2006.

\bibitem{de2005learning}
Ernesto De~Vito, Lorenzo Rosasco, Andrea Caponnetto, Umberto De~Giovannini, and
  Francesca Odone.
\newblock Learning from examples as an inverse problem.
\newblock {\em Journal of Machine Learning Research}, 6(5), 2005.

\bibitem{devito2014}
Ernesto {De Vito}, Lorenzo Rosasco, and Alessandro Toigo.
\newblock Learning sets with separating kernels.
\newblock {\em Applied and Computational Harmonic Analysis}, 37(2):185--217,
  2014.

\bibitem{dinu2022aggregation}
Marius-Constantin Dinu, Markus Holzleitner, Maximilian Beck, Hoan~Duc Nguyen,
  Andrea Huber, Hamid Eghbal-zadeh, Bernhard~A Moser, Sergei Pereverzyev, Sepp
  Hochreiter, and Werner Zellinger.
\newblock Addressing parameter choice issues in unsupervised domain adaptation
  by aggregation.
\newblock {\em International Conference on Learning Representations}, 2023.

\bibitem{evgeniou2000}
Theodoros Evgeniou, Massimiliano Pontil, and Tomaso~A. Poggio.
\newblock Regularization networks and support vector machines.
\newblock {\em Advances in Computational Mathematics}, 13(1):1--50, 2000.

\bibitem{gizewski2022regularization}
Elke~R Gizewski, Lukas Mayer, Bernhard~A Moser, Duc~Hoan Nguyen, Sergiy
  Pereverzyev~Jr, Sergei~V Pereverzyev, Natalia Shepeleva, and Werner
  Zellinger.
\newblock On a regularization of unsupervised domain adaptation in rkhs.
\newblock {\em Applied and Computational Harmonic Analysis}, 57:201--227, 2022.

\bibitem{hido2011statistical}
Shohei Hido, Yuta Tsuboi, Hisashi Kashima, Masashi Sugiyama, and Takafumi
  Kanamori.
\newblock Statistical outlier detection using direct density ratio estimation.
\newblock {\em Knowledge and information systems}, 26:309--336, 2011.

\bibitem{huang2006}
Jiayuan Huang, Alexander~J. Smola, Arthur Gretton, Karsten~M. Borgwardt, and
  B.~Sch{\"o}lkopf.
\newblock Correcting sample selection bias by unlabeled data.
\newblock {\em Advances in neural information processing systems}, 19:601--608,
  2006.

\bibitem{kanamori2009}
Takafumi Kanamori, Shohei Hido, and Masashi Sugiyama.
\newblock A least-squares approach to direct importance estimation.
\newblock {\em Journal of Machine Learning Research}, 10:1391--1445, 07 2009.

\bibitem{kanamori2011f}
Takafumi Kanamori, Taiji Suzuki, and Masashi Sugiyama.
\newblock $ f $-divergence estimation and two-sample homogeneity test under
  semiparametric density-ratio models.
\newblock {\em IEEE Transactions on Information Theory}, 58(2):708--720, 2011.

\bibitem{kanamori2012}
Takafumi Kanamori, Taiji Suzuki, and Masashi Sugiyama.
\newblock Statistical analysis of kernel-based least-squares density-ratio
  estimation.
\newblock {\em Machine Learning}, 86:335--367, 2012.

\bibitem{kato2019learning}
Masahiro Kato, Takeshi Teshima, and Junya Honda.
\newblock Learning from positive and unlabeled data with a selection bias.
\newblock In {\em International conference on learning representations}, 2019.

\bibitem{keziou2005test}
Amor Keziou and Samuela Leoni-Aubin.
\newblock Test of homogeneity in semiparametric two-sample density ratio
  models.
\newblock {\em Comptes Rendus Math{\'e}matique}, 340(12):905--910, 2005.

\bibitem{shuai2019Analysis}
Shuai Lu, Peter Math\'{e}, and Sergiy Pereverzyev.
\newblock Analysis of regularized {N}ystr{\"o}m subsampling for regression
  functions of low smoothness.
\newblock {\em Analysis and Applications}, 17(06):931--946, 2019.

\bibitem{lu2018}
Shuai Lu, Peter Mathé, and Sergei Pereverzyev.
\newblock Balancing principle in supervised learning for a general
  regularization scheme.
\newblock {\em Applied and Computational Harmonic Analysis}, 48, 2018.

\bibitem{1stbook}
Shuai Lu and Sergei~V. Pereverzev.
\newblock {\em Regularization Theory for Ill-posed Problems - Selected Topics}.
\newblock De Gruyter, Berlin, Boston, 2013.

\bibitem{mathe2008}
Peter Mathe and Bernd Hofmann.
\newblock How general are general source conditions?
\newblock {\em Inverse Problems}, 24(1):015009, jan 2008.

\bibitem{meyer2021domain}
Anneke Meyer, Alireza Mehrtash, Marko Rak, Oleksii Bashkanov, Bjoern Langbein,
  Alireza Ziaei, Adam~S Kibel, Clare~M Tempany, Christian Hansen, and Junichi
  Tokuda.
\newblock Domain adaptation for segmentation of critical structures for
  prostate cancer therapy.
\newblock {\em Scientific Reports}, 11(1):1--14, 2021.

\bibitem{mohamed2016learning}
Shakir Mohamed and Balaji Lakshminarayanan.
\newblock Learning in implicit generative models.
\newblock {\em arXiv preprint arXiv:1610.03483}, 2016.

\bibitem{nguyen2023regularized}
DH~Nguyen, W~Zellinger, and S~Pereverzyev.
\newblock On regularized radon-nikodym differentiation.
\newblock {\em RICAM-Report 2023-13}, 2023.

\bibitem{nguyen2007estimating}
XuanLong Nguyen, Martin~J Wainwright, and Michael Jordan.
\newblock Estimating divergence functionals and the likelihood ratio by
  penalized convex risk minimization.
\newblock {\em Advances in neural information processing systems}, 20, 2007.

\bibitem{nguyen2010}
XuanLong Nguyen, Martin~J. Wainwright, and Michael~I. Jordan.
\newblock Estimating divergence functionals and the likelihood ratio by convex
  risk minimization.
\newblock {\em IEEE Transactions on Information Theory}, 56(11):5847--5861,
  2010.

\bibitem{nikzad2018domain}
Ramin Nikzad-Langerodi, Werner Zellinger, Edwin Lughofer, and Susanne
  Saminger-Platz.
\newblock Domain-invariant partial-least-squares regression.
\newblock {\em Analytical Chemistry}, 90(11):6693--6701, 2018.

\bibitem{oneto2016}
Luca Oneto, Sandro Ridella, and Davide Anguita.
\newblock Tikhonov, {I}vanov and {M}orozov regularization for support vector
  machine learning.
\newblock {\em Machine Learning}, 103(1):103--136, 2016.

\bibitem{page2019}
Stephen Page and Steffen Gr{\"u}new{\"a}lder.
\newblock Ivanov-regularised least-squares estimators over large rkhss and
  their interpolation spaces.
\newblock {\em J. Mach. Learn. Res.}, 20(120):1--49, 2019.

\bibitem{pauwels2018}
Edouard Pauwels, Francis Bach, and Jean-Philippe Vert.
\newblock Relating leverage scores and density using regularized christoffel
  functions.
\newblock In S.~Bengio, H.~Wallach, H.~Larochelle, K.~Grauman, N.~Cesa-Bianchi,
  and R.~Garnett, editors, {\em Advances in Neural Information Processing
  Systems}, volume~31. Curran Associates, Inc., 2018.

\bibitem{que2013}
Qichao Que and Mikhail Belkin.
\newblock Inverse density as an inverse problem: The {F}redholm equation
  approach.
\newblock {\em Advances in Neural Information Processing Systems}, 26:., 2013.

\bibitem{rosasco2008}
Lorenzo Rosasco, Francesca Odone, Ernesto De~Vito, and Alessandro Verri.
\newblock Spectral algorithms for supervised learning.
\newblock {\em Neural computation}, 20, 08 2008.

\bibitem{rudi2015}
Alessandro Rudi, Raffaello Camoriano, and Lorenzo Rosasco.
\newblock Less is more: Nystr\"{o}m computational regularization.
\newblock In C.~Cortes, N.~Lawrence, D.~Lee, M.~Sugiyama, and R.~Garnett,
  editors, {\em Advances in Neural Information Processing Systems}, volume~28.
  Curran Associates, Inc., 2015.

\bibitem{schuster2020}
Ingmar Schuster, Mattes Mollenhauer, Stefan Klus, and Krikamol Muandet.
\newblock Kernel conditional density operators.
\newblock In Silvia Chiappa and Roberto Calandra, editors, {\em Proceedings of
  the Twenty Third International Conference on Artificial Intelligence and
  Statistics}, volume 108 of {\em Proceedings of Machine Learning Research},
  pages 993--1004. PMLR, 26--28 Aug 2020.

\bibitem{shimodaira2000}
Hidetoshi Shimodaira.
\newblock Improving predictive inference under covariate shift by weighting the
  log-likelihood function.
\newblock {\em Journal of Statistical Planning and Inference}, 90:227--244, 10
  2000.

\bibitem{smale2007}
Steve Smale and Ding-Xuan Zhou.
\newblock Learning theory estimates via integral operators and their
  approximations.
\newblock {\em Constructive Approximation}, 26:153--172, 2007.

\bibitem{smola2009relative}
Alex Smola, Le~Song, and Choon~Hui Teo.
\newblock Relative novelty detection.
\newblock In {\em Artificial Intelligence and Statistics}, pages 536--543.
  PMLR, 2009.

\bibitem{sugiyama2005}
Masashi Sugiyama and Klaus-Robert Müller.
\newblock Input-dependent estimation of generalization error under covariate
  shift.
\newblock {\em Statistics \& Decisions}, 23:249--279, 01 2005.

\bibitem{sugiyama2012book}
Masashi Sugiyama, Taiji Suzuki, and Takafumi Kanamori.
\newblock {\em Density Ratio Estimation in Machine Learning}.
\newblock Cambridge University Press, 2012.

\bibitem{vapnik2013nature}
Vladimir~N Vapnik.
\newblock {\em The Nature of Statistical Learning Theory}.
\newblock Springer science \& business media, 2013.

\end{thebibliography}

\end{document}